\documentclass{article}

\usepackage[margin=1in]{geometry}
\usepackage[numbers]{natbib}

\usepackage[utf8]{inputenc}  
\usepackage[T1]{fontenc}     
\usepackage{hyperref}        
\usepackage{url}             
\usepackage{amsfonts}        
\usepackage{amsmath}
\usepackage{nicefrac}        
\usepackage{microtype}       
\usepackage[table]{xcolor}   
\usepackage{ifthen}

\usepackage{array}
\usepackage[caption=false,font=normalsize,labelfont=sf,textfont=sf]{subfig}
\usepackage{textcomp}
\usepackage{stfloats}
\usepackage{graphicx}
\usepackage{multirow}
\usepackage{diagbox}
\usepackage{booktabs}        
\usepackage{float}
\usepackage[ruled,vlined]{algorithm2e}
\usepackage{tikz}
\usetikzlibrary{shapes.geometric, arrows.meta, positioning, fit, backgrounds, calc}
\usepackage{xintexpr}
\usepackage{xstring}

\renewcommand{\paragraph}[1]{\textbf{\StrGobbleRight{#1}{1}:}}

\definecolor{darkgreen}{RGB}{0,100,0}

\newcommand{\res}[3]{\ensuremath{%
    \ifthenelse{\equal{#1}{1}}%
    {
        \mathbf{#2}\mathbin{\textcolor{gray}{\pm}}\textcolor{gray}{\xintifboolexpr{#3<10}{0}{2}#3}%
    }%
    {
        #2\mathbin{\textcolor{gray}{\pm}}\textcolor{gray}{\xintifboolexpr{#3<10}{0}{}#3}%
    }%
}}

\title{LM-GRASP: Instance-Specific Language Models for Combinatorial Construction via Online Imitation Learning}

\date{}

\author{
Mohand Mezmaz \\
SnT, University of Luxembourg \\
\texttt{mohand.mezmaz@uni.lu}
\and
Grégoire Danoy \\
SnT, University of Luxembourg \\
\texttt{gregoire.danoy@uni.lu}
}

\begin{document}

\maketitle

\begin{abstract}
Machine learning has increasingly been integrated into
combinatorial optimization, most commonly through neural constructors
trained via reinforcement learning on large offline datasets for a
fixed problem class. Such approaches incur substantial pretraining
costs and often generalize poorly outside their training distribution.
We propose a different paradigm: rather than learning a constructor
offline, we introduce a metaheuristic framework that reformulates the
randomized constructive phase of the Greedy Randomized Adaptive Search
Procedure (GRASP) as an online imitation learning task, learned
from scratch on each problem instance. Under our paradigm, a local
search procedure serves as an expert oracle, while a decoder-only
Transformer network acts as the constructive policy. Classical GRASP
variants rely on static, hand-crafted heuristic rules that are
inherently myopic, ranking candidates via a localized scalar cost and
unable to exploit structural regularities discovered elsewhere during
the search. In contrast, our approach is entirely data-driven. The
sequential construction policy emerges directly from high-quality
solutions discovered during the optimization process, requiring no
explicit encoding of problem-specific features.

We instantiate our methodology as \textbf{LM-GRASP}, a hybrid
metaheuristic operating via an iterative learn--infer--improve cycle.
Our framework trains the autoregressive policy online via behavioral cloning on
a dynamic archive of elite solution trajectories, using no
external data or offline pretraining. The underlying
pipeline interfaces with the combinatorial domain solely through the
objective evaluator used by the local search. Consequently, the system requires no
domain-specific heuristic design.

We evaluate LM-GRASP on the Permutation Flow-Shop Scheduling Problem
(PFSP) under the makespan minimization objective. We test our model using the standard
Taillard benchmark suite (\texttt{ta51}--\texttt{ta60}). The selected block 
is highly challenging because half of the optimal solutions remain
unknown, making the configuration the most discriminating setup for
approximate methods. On identical hardware,
LM-GRASP outperforms GPU-GRASP by $28.4$ makespan units on average.
Our algorithmic gain is similar in magnitude to the improvement brought by GPU 
acceleration over sequential execution ($27.2$ units), though with 
overlapping standard deviations. These results suggest that instance-specific,
online-trained language models represent a promising and practical
alternative to hand-engineered constructors in metaheuristic search. 
The performance edge holds particularly true for combinatorial problems whose solution 
landscape resists classical greedy construction.
\end{abstract}

\section{Introduction}
\label{sec:introduction}

Combinatorial optimization problems arise in a broad range of scientific
and industrial domains, including scheduling, logistics, network design,
and resource allocation. These problems are characterized by a discrete,
exponentially large solution space, a trait that renders exact methods
intractable for large-scale instances. Metaheuristics such as the Greedy
Randomized Adaptive Search Procedure (GRASP)~\cite{feo1995} are therefore
widely employed to obtain high-quality approximate solutions within a
bounded computational budget. Over the years, numerous GRASP
variants and hybrids have been proposed to improve upon the basic
scheme, including path-relinking and parallel hybridizations that
combine GRASP with other metaheuristic components~\cite{ravetti2012},
and hyper-heuristic frameworks that coordinate large pools of
hand-tuned construction configurations~\cite{alekseeva2017}. These
hybrids, however, remain fundamentally reliant on hand-engineered
construction or selection rules, leaving the core limitation of static,
human-designed heuristics unaddressed.

In parallel, machine learning has increasingly been integrated
into metaheuristic and combinatorial optimization pipelines, from
neural constructors trained via reinforcement learning for routing
problems~\cite{kool2019,kwon2020,chin2024} to large language models
used as configuration agents or heuristic-code generators for
metaheuristics~\cite{martinek2024,liu2024}. These approaches demonstrate
the potential of learned components to reduce the manual engineering
burden of metaheuristic design, but they typically require costly
offline pretraining on large problem-class datasets, or rely on
general-purpose models that are not adapted to the specific instance
being solved. The integration of language models directly into the
search loop of a classical metaheuristic such as GRASP, trained online
and without external data, remains comparatively underexplored.

A central component of GRASP is its randomized constructive phase, in
which a feasible solution is built incrementally. At each step, a
candidate element is drawn stochastically from a Restricted Candidate
List (RCL). The generated list contains the most promising options evaluated according to a
local cost metric. The resulting construction rule is inherently
\emph{human-engineered}. It encodes domain knowledge explicitly, demands
extensive problem-specific tuning, and remains invariant throughout the
entire search horizon. Optimization efficacy is constrained by the
myopic nature of the greedy criterion. The core greedy metric evaluates decisions through
a localized snapshot of the partial sequence, failing to exploit global
structural patterns. Efficacy is also limited by the quality ceiling imposed by the
hand-crafted cost function itself.

To address these limitations, we propose a different paradigm. Rather
than engineering a construction rule a priori, we \emph{learn} the rule from
high-quality solutions discovered during the optimization process. Our framework
reformulates the traditional constructive phase as an online imitation
learning task~\cite{pomerleau1991}. In our mapping, the local search
procedure acts as an inductive \emph{expert oracle} that refines
stochastically generated candidates into local optima. A decoder-only
Transformer serves as the generative \emph{parametric policy}. Finally, an
active archive of the refined solutions forms the
\emph{demonstration corpus} of expert-endorsed trajectories.

A key operational property of our framework is that the sequential
construction policy requires \emph{no explicit encoding of problem
topology}. The pipeline interfaces with the combinatorial domain solely
through the objective function evaluator used by the local search. As a result,
the constructive logic emerges entirely from data-driven sequence
modeling. By framing combinatorial construction as sequential token
generation, decoder-only Transformer architectures can capture
non-myopic dependencies across the partial solution. The network conditions each
constructive decision on the full history of preceding choices. This approach contrasts
with classical greedy criteria that rely strictly on local scalar metrics. The details
of the policy mapping and the iterative optimization loop are formalized in
Section~\ref{sec:method}.

We instantiate our methodology as \textbf{LM-GRASP}, a hybrid
metaheuristic that replaces the hand-crafted constructive phase of GRASP
with an instance-specific, online-trained language model. The policy is
optimized via behavioral cloning on the elite solutions accumulated
during search, and the parameters are iteratively retrained as the archive evolves.
LM-GRASP is not proposed as a universal alternative to GRASP. Instead, our framework serves as a
targeted complement suited to problem regimes of sufficient complexity,
where the overhead of online policy training is offset by the quality gains
of a learned constructor.

We evaluate LM-GRASP on the Permutation Flow-Shop Scheduling Problem
(PFSP) under the makespan minimization objective using the Taillard
benchmark suite (\texttt{ta51}--\texttt{ta60}). The $50 \times 20$
block is the first configuration in the benchmark for which half of the
absolute optima remain unknown, positioning our experiments at the
genuine complexity frontier where the structural advantage of a learned
policy over a classical greedy heuristic can be most directly assessed.
Under a fixed 5-hour wall-clock budget, LM-GRASP consistently surpasses
both the sequential CPU-GRASP and GPU-GRASP baselines. Our
three-tiered hardware setup decouples two performance vectors: an
algorithmic learning delta of $28.4$ units (GPU-GRASP vs.\ LM-GRASP)
and a hardware acceleration delta of $27.2$ units (CPU-GRASP vs.\
GPU-GRASP). The two deltas are of comparable magnitude, suggesting that
a data-driven autoregressive policy can yield optimization gains on par
with porting the classical heuristic to a massively parallelized GPU.

The primary contributions of this work are as follows:
\begin{itemize}
    \item \textbf{A Paradigm Reformulation (\S\ref{sec:method}):} We
    present a formal framework mapping greedy randomized combinatorial
    construction to an online imitation learning task, with local search
    as an inductive expert oracle and a decoder-only Transformer as the
    constructive policy.

    \item \textbf{The LM-GRASP Framework (\S\ref{sec:method}):} We
    develop an instance-specific, data-driven hybrid metaheuristic that
    replaces hand-engineered constructive rules with a policy learned
    entirely online. Our approach requires no external training data or offline
    pretraining.

    \item \textbf{Algorithmic vs.\ Hardware Quantification
    (\S\ref{sec:experiments}):} We provide empirical evidence on Taillard
    benchmarks that LM-GRASP consistently outperforms both baselines. We 
    also isolate the algorithmic contribution from the hardware contribution
    via a controlled three-tiered experimental design.

    \item \textbf{Instance-Complexity Positioning
    (\S\ref{sec:background}, \S\ref{sec:experiments}):} We characterize
    the complexity frontier of the Taillard benchmark and identify the
    $50 \times 20$ block as the principled evaluation setting. At the selected frontier,
    a learned construction policy becomes competitive with, and subsequently 
    outperforms, both classical construction and hardware acceleration.

    \item \textbf{Taxonomy and Future Outlook (\S\ref{sec:related},
    \S\ref{sec:conclusion}):} We contribute a structured taxonomy of
    language model-integrated optimization approaches, positioning
    LM-GRASP within the research landscape. Finally, we identify cross-domain validation,
    continuous policy updates, and hyperparameter optimization as
    the primary directions for future work.
\end{itemize}

The remainder of this paper is organized as follows.
Section~\ref{sec:background} establishes the theoretical background,
outlines classical GRASP mechanics, and motivates the choice of the PFSP
as the evaluation domain. Section~\ref{sec:related} reviews
machine learning-driven optimization and situates LM-GRASP within a
structured taxonomy. Section~\ref{sec:method} details the architectural
and algorithmic design of LM-GRASP. Section~\ref{sec:experiments}
presents the experimental configuration and empirical results.
Section~\ref{sec:conclusion} concludes with a summary of findings and
directions for future research.

\section{Background}
\label{sec:background}

This section establishes the theoretical and algorithmic foundations
underlying LM-GRASP. We begin by formalizing the Permutation Flow-Shop
Scheduling Problem (PFSP) and its classical metaheuristic solution
framework, GRASP. We then introduce the two learning-based components
that replace the hand-crafted constructive phase. The components consist of imitation learning via
behavioral cloning, alongside autoregressive sequence generation via
decoder-only Transformer architectures. We conclude by positioning
LM-GRASP relative to classical GRASP and motivating the choice of the
PFSP as our primary evaluation domain.

\subsection{The Permutation Flow-Shop Scheduling Problem}

Given a set of $n$ jobs and $m$ machines arranged in a fixed processing
sequence, each job $j$ must be processed on every machine $i$. Every job requires a
strictly positive duration $p_{ij} > 0$. The operational constraints
require that all machines process jobs in the same order. The rule defines the 
\emph{permutation} flow-shop setting. Furthermore, preemption is prohibited, and
each machine can process at most one job at a time.

A feasible solution is defined by a permutation $\pi = (\pi_1, \dots,
\pi_n)$ of the $n$ jobs. The completion time $C_i(\pi, k)$ of the job
at position $k$ on machine $i$ satisfies the following recurrence:
\begin{align}
C_1(\pi, 1) &= p_{1,\pi_1}, \\
C_i(\pi, 1) &= C_{i-1}(\pi, 1) + p_{i,\pi_1}, \quad \text{for } i > 1, \\
C_1(\pi, k) &= C_1(\pi, k-1) + p_{1,\pi_k}, \quad \text{for } k > 1, \\
C_i(\pi, k) &= \max\!\bigl(C_{i-1}(\pi, k),\; C_i(\pi, k-1)\bigr) +
p_{i,\pi_k}, \quad \text{for } i > 1, \, k > 1.
\end{align}

The objective is to find the permutation $\pi^*$ that minimizes the
\emph{makespan} $C_{\max}(\pi) = C_m(\pi, n)$. The target value represents the completion time of
the final job on the final machine. PFSP makespan minimization is strongly
NP-hard for $m \geq 3$~\cite{garey1976}.

For empirical evaluation, we use the ten Taillard instances
\texttt{ta51}--\texttt{ta60}~\cite{taillard1993}. Each benchmark contains 
$n = 50$ jobs and $m = 20$ machines. The selected instances represent a standard benchmark 
in the scheduling literature with established best-known makespans.

Our choice of the PFSP as the evaluation domain is motivated by two
structural arguments. First, the PFSP operates over a \emph{pure
permutation space}: unlike problems such as the Knapsack Problem, where
flipping a single binary variable $x_i \in \{0,1\}$ typically preserves
feasibility, any incorrect token assignment in a permutation
immediately destroys the validity of the entire sequence, making the
PFSP a stringent test for autoregressive generators, which must
maintain global feasibility throughout the construction trajectory.

Second, the PFSP exhibits higher \emph{representational complexity}
than other permutation problems of comparable size. In the Euclidean
TSP, each city is fully described by a two-dimensional coordinate
vector, a compact representation that admits efficient exact methods
(a 20-city instance is solvable in minutes by the Held-Karp
algorithm~\cite{held1962dynamic}). In the PFSP, each job is instead
characterized by an $m$-dimensional vector of processing times, one per
machine, so difficulty scales directly with $m$: a $20 \times 20$
instance requires days of Branch-and-Bound
computation~\cite{mezmaz2007grid}, whereas a TSP instance of the same
job count is solved trivially. This dependence on $m$ makes the
20-machine PFSP a demanding testbed for a language model's capacity to
capture structural dependencies among solution components.

\subsection{GRASP}

The Greedy Randomized Adaptive Search Procedure
(GRASP)~\cite{feo1995} is a multi-start iterative metaheuristic. Each
iteration consists of two distinct phases. The process begins with a \emph{greedy randomized
construction} that produces a feasible solution, which is followed by a
\emph{local search} that refines the output to a local optimum.
Algorithm~\ref{alg:grasp} details the general procedure.

\begin{algorithm}[H]
\caption{GRASP}
\label{alg:grasp}
\KwIn{Cost function $f$, greedy parameter $\alpha \in [0,1]$,
      neighborhood $\mathcal{N}$, time budget $T$}
\KwOut{Best solution found $x^*$}
$x^* \leftarrow \textsc{null}$\;
\While{elapsed time $< T$}{
    $x \leftarrow \textsc{GreedyRandomizedConstruction}(f, \alpha)$\;
    $x \leftarrow \textsc{LocalSearch}(x, \mathcal{N})$\;
    \If{$f(x) < f(x^*)$}{$x^* \leftarrow x$}
}
\Return $x^*$\;
\end{algorithm}

\paragraph{Greedy Randomized Construction.}
Starting from an empty partial sequence, the heuristic evaluates each
available candidate element $e$ using an incremental cost metric
$c(e)$. The algorithm computes the bounds $[c_{\min}, c_{\max}]$ and builds a
Restricted Candidate List:
$$\text{RCL}(\alpha) = \{e \mid c(e) \leq c_{\min} + \alpha(c_{\max} - c_{\min})\}$$
One element is sampled uniformly at random from the filtered list and appended to the sequence. 
The parameter $\alpha$ interpolates between a purely greedy heuristic ($\alpha = 0$)
and a uniform random walk ($\alpha = 1$).

For the PFSP, the incremental cost of inserting job $j$ at position $k$
is typically the makespan $C_{\max}(\pi^k)$ of the resulting partial
schedule~\cite{taillard1993}. The objective function is entirely hand-crafted. 
The heuristic embeds domain-specific knowledge and requires a complete redesign when applied to
a different combinatorial problem.

\paragraph{Limitations of Classical Construction.}
The classical GRASP constructive phase suffers from two structural limitations.
First, the process is \emph{myopic}. Candidate ranking relies entirely on a scalar,
localized incremental cost. The metric cannot exploit global structural patterns
in the partial solution. Second, the constructor is \emph{memoryless}. Each
construction trajectory is generated independently. As a result, the algorithm
cannot reuse structural regularities from previously discovered,
high-quality solutions. Together, the limitations impose a quality
ceiling that additional hardware parallelism alone cannot overcome.

\subsection{Imitation Learning and Behavioral Cloning}

Imitation learning~\cite{hussein2017} trains a parametric policy
$\pi_\theta$ to replicate the sequential decision-making behavior of an
expert oracle $\pi^*$. This optimization utilizes a corpus of demonstrated trajectories.
\textbf{Behavioral Cloning} (BC)~\cite{pomerleau1991} is the simplest
instantiation of the paradigm. The technique reduces imitation to supervised learning by maximizing
the log-likelihood of expert actions over a dataset of state-action
pairs:
$$ \mathcal{L}_{\text{BC}}(\theta) = -\mathbb{E}_{(s,a) \sim \mathcal{D}} \left[ \log \pi_\theta(a \mid s) \right] $$
where $\mathcal{D}$ is the demonstration dataset, $s$ is the current
state, and $a$ is the corresponding expert action.

In our setting, states correspond to partial permutations, while actions correspond to
the placement of the next job. The expert oracle is instantiated by
the local search procedure. Each demonstration is therefore a locally optimal
permutation, representing the expert's response to the constructive task. The
policy $\pi_\theta$ is trained to replicate the token-by-token
construction logic implicit within the elite trajectories.

\subsection{Transformer Architecture and Decoder-only Autoregressive Generation}

We parameterize the constructive policy $\pi_\theta$ with a
decoder-only Transformer~\cite{vaswani2017}: an $L$-layer stack of
\emph{masked causal self-attention} blocks, using learned positional
embeddings, that models a probability distribution over discrete
sequences. Given a partial sequence $x_{1:t}$, the model predicts the
next token $x_{t+1}$, with each position attending only to preceding
tokens, so that predictions are conditioned on the full construction
history rather than on a local scalar summary.

Autoregressive inference proceeds sequentially. Starting from a
Begin-of-Sequence (\textsc{bos}) token, the model samples $x_{t+1}
\sim p_\theta(\cdot \mid x_{1:t})$ and appends the selection to the current
context. The process repeats until all combinatorial constraints are satisfied
or an End-of-Sequence (\textsc{eos}) token is produced. Generation
diversity is controlled by a softmax temperature $\tau > 0$ and top-$k$
truncation, providing a direct mechanism to balance exploration and
exploitation.

Model parameters are optimized by minimizing the causal cross-entropy
loss:
$$ \mathcal{L}(\theta) = -\frac{1}{n}\sum_{t=1}^{n} \log p_\theta(x_t \mid x_{1:t-1}) $$
The objective is equivalent to the behavioral cloning loss
$\mathcal{L}_{\text{BC}}$ applied directly to sequential decisions. Training the
Transformer on elite local optima encourages the policy to
internalize long-range structural dependencies across the solution. A myopic, 
scalar-driven greedy heuristic cannot replicate our architecture's structural capacity.

\section{Related Work}
\label{sec:related}

This section situates LM-GRASP within the landscape of machine learning
for combinatorial optimization. We first provide a brief overview
of learning-assisted combinatorial optimization, covering both offline
neural constructors and foundation models, and imitation learning for
algorithmic sub-routines. We then
propose a structural taxonomy to classify language model-based solvers
and position LM-GRASP within the unified matrix.

\subsection{Learning-Assisted Combinatorial Optimization}

The use of neural networks to assist combinatorial optimization
predates the recent deep learning era, with early work exploring
Hopfield networks for relaxed combinatorial decision problems~\cite{hopfield1985}.
The field has since grown substantially, and several recent surveys
provide a comprehensive overview of learning-based approaches to
combinatorial optimization, ranging from supervised and reinforcement
learning constructors to the more recent integration of large language
models into the optimization pipeline~\cite{daros2025}.

The intersection of machine learning and combinatorial optimization has
grown rapidly since Pointer Networks~\cite{vinyals2015} first
demonstrated autoregressive construction of TSP solutions. Kool et
al.~\cite{kool2019} integrated the routing template with the Transformer
architecture. Their study established a line of deep reinforcement learning (DRL)
frameworks for routing problems. POMO~\cite{kwon2020} extended the
paradigm via multiple rollouts from diverse starting nodes, achieving
competitive results on the TSP and CVRP. However, reinforcement frameworks suffer from a shared limitation. A separate model must be trained for each problem
size and configuration, making the offline training cost substantial.

More recently, Chin et al.~\cite{chin2024} addressed cross-instance
generalization with FM-MCVRP. The framework utilizes a foundation model pretrained on 38.1
million problem--solution pairs for the capacitated vehicle routing
problem. While the platform represents a highly data-intensive instantiation of
a learned constructor, the pipeline requires access to a large offline
dataset. In contrast, LM-GRASP requires no external data. The training corpus is
synthesized online using solutions discovered for the specific instance at hand.
Across this trajectory, the dominant pattern is a clear trade-off:
stronger generalization across instances is purchased at the cost of an
increasingly large and expensive offline training corpus, from
single-problem-size DRL policies~\cite{kool2019} to foundation models
pretrained on tens of millions of solved instances~\cite{chin2024}.

A second, more narrowly scoped trajectory uses imitation learning to
approximate specific algorithmic sub-routines rather than end-to-end
solution construction -- for example, branching strategies in exact
solvers~\cite{gasse2019} or local rewriting operators~\cite{chen2019}.
Both lines of work train their imitation policy offline, once, on a
fixed dataset of expert demonstrations collected prior to deployment.
Behavioral cloning has
also been studied more broadly as a general-purpose imitation learning
technique~\cite{pomerleau1991}, with known limitations such as
distribution shift between the training and inference state
distributions, motivating extensions such as DAgger-style trajectory
aggregation~\cite{ross2011}.
LM-GRASP shares the philosophy of
replacing hand-crafted logic with learned approximations. However, our architecture targets
the randomized construction phase of metaheuristics rather than
tree-search navigation. The system also trains exclusively on dynamically generated
demonstrations instead of static offline datasets. To our
knowledge, no prior work applies imitation learning \emph{online}, with
the demonstration corpus generated and refreshed continuously during
the search for a single target instance, rather than collected once
offline.

\subsection{A Taxonomy of Language Model-Based Optimization}

\begin{figure}[H]
\centering
\includegraphics[width=0.82\textwidth]{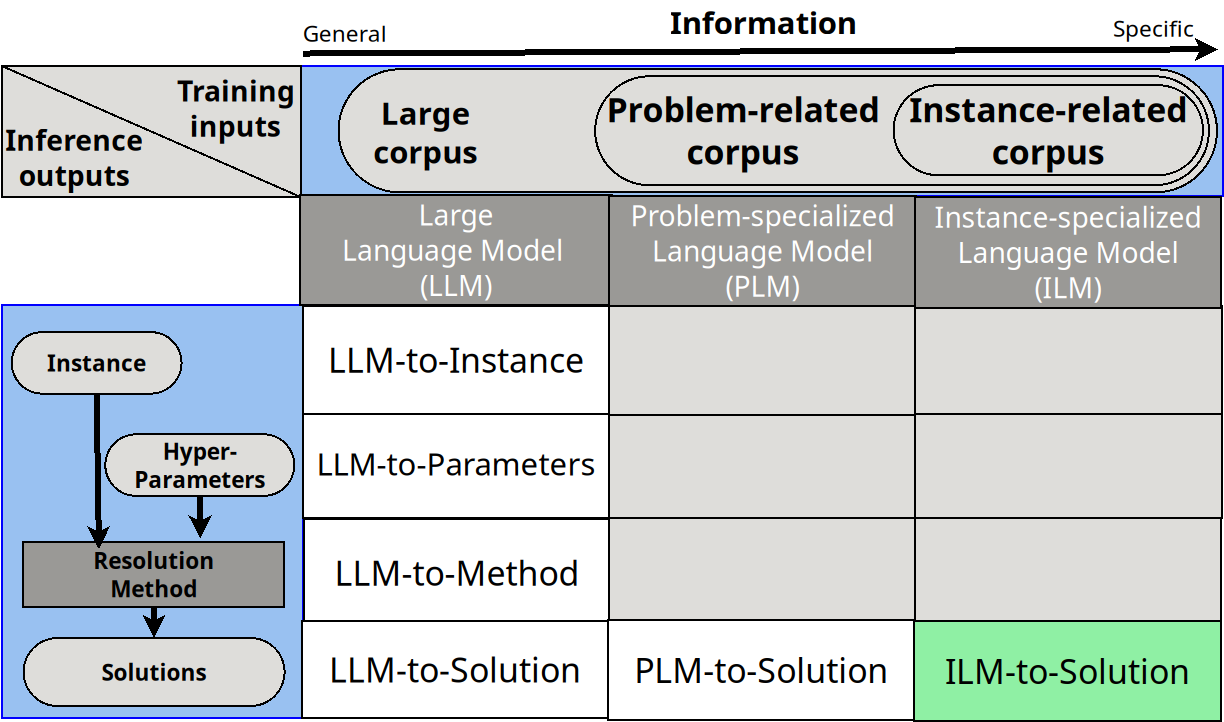}
\caption{Taxonomy of language model-based optimization approaches along
two axes: training corpus specialization (columns: generic LM,
problem-class PLM, instance-specific ILM) and functional role at
inference (rows: instance, parameters, method, solution). The
cell highlighted in green (ILM-to-Solution) denotes the
configuration introduced in this work: to our knowledge, LM-GRASP is
the first method to occupy this cell, training a generative
constructor entirely online and from scratch on instance-specific
demonstrations, with no external data or offline pretraining.}
\label{fig:taxonomy}
\end{figure}

To organize the growing landscape of language model-integrated
optimization, we propose a taxonomy along two axes, illustrated in
Figure~\ref{fig:taxonomy}:

\begin{itemize}
    \item \textbf{Training Corpus Specialization:} This axis scales from \emph{generic}
    approaches (a general-purpose LLM pretrained on natural language), to
    \emph{problem-class-specific} setups (a model trained offline on solutions
    from a fixed problem class), and finally to \emph{instance-specific} methods (a model
    trained online on solutions generated strictly for the target instance).

    \item \textbf{Functional Role at Inference:} This axis dictates the generative output
    of the model. Outputs can include a \emph{problem instance} (such as algebraic
    constraints for a solver), a \emph{resolution method} (such as
    executable heuristic code or search configurations), or a direct
    \emph{solution candidate} (such as a permutation, tour, or assignment
    vector).
\end{itemize}

Crossing these two axes yields twelve possible categories, of
which six are populated by existing or proposed work; we describe each
below, along with representative examples and brief remarks on their
relative maturity.

\textbf{LLM-to-Instance:} The model translates a natural
language problem description into a formal mathematical specification
(e.g., a linear program), which is then delegated to a classical
solver~\cite{huang2025}. This category is appealing as an accessibility
layer over existing solvers, since it requires no change to the
underlying optimization machinery, but its output quality is bounded by
the correctness of the generated formalization rather than by any
search capability of the model itself.

\textbf{LLM-to-Parameters:} The model acts as a configuration
agent, tuning hyperparameters such as population size, mutation rate,
or crossover rate without modifying the underlying search
logic~\cite{martinek2024}. This is among the lowest-risk integrations
of LLMs into metaheuristics, since the model never directly touches
solution quality, only the algorithm's operating regime; it is
correspondingly one of the more mature and widely explored categories.

\textbf{LLM-to-Method:} The model synthesizes executable
heuristic code or algorithmic structures that are compiled and
run~\cite{martinek2024,liu2024}. Liu et al.~\cite{liu2024} represent a
prominent instantiation of this category, evolving heuristic source code
through LLM-guided mutation. This category effectively delegates
algorithm design itself to the language model, and has rapidly become
one of the most active areas of LLM-for-optimization research, owing to
its compatibility with general-purpose, off-the-shelf LLMs.

\textbf{LLM-to-Solution:} The model directly outputs a
solution candidate. Elhenawy et al.~\cite{elhenawy2024} demonstrate
the methodology for the TSP, prompting a general-purpose
multimodal LLM with visual and textual instance representations, while Zhang et al.~\cite{zhang2024} document the 
underlying scalability limitations of small-scale, general-purpose
LLMs used as direct numerical optimizers. This category is, to date,
the least reliable in the taxonomy: because the model receives no
problem-specific training signal, solution quality degrades sharply as
instance size grows, and feasibility itself is not guaranteed.

\textbf{PLM-to-Solution:} A problem-class-specific model
    trained offline on domain-specific data~\cite{kool2019,kwon2020,
    chin2024}. The targeted systems achieve strong inference performance but carry
    high pretraining costs and generalize poorly across different problem classes.
    This is the most densely populated category in the taxonomy
    and the one with the longest research history, tracing back to
    Pointer Networks~\cite{vinyals2015}; it is also the category for
    which empirical performance is best documented, owing to its
    maturity and the relative ease of benchmarking a fixed, pretrained
    policy.

\textbf{ILM-to-Solution:} An instance-specific model trained
    online to generate solution candidates directly. \textbf{LM-GRASP
    occupies the designated cell}. The Transformer is trained from scratch on
    elite solutions discovered for the active instance, requiring no
    external data or pretraining. To our knowledge, this is the
    first instantiation of this category: while instance-specific
    \emph{conditioning} has been explored within problem-class-specific
    architectures (e.g., fine-tuning a pretrained policy on a single
    instance), no prior approach trains a generative solution
    constructor entirely from scratch, online, using only solutions
    discovered for the target instance, without recourse to any
    external dataset or pretrained checkpoint.

\paragraph{The gap addressed by LM-GRASP.}
The taxonomy above reveals a structural gap in the existing landscape.
Methods that achieve strong, well-documented solution quality
(PLM-to-Solution) do so only by amortizing a substantial offline
training cost over a fixed problem class, and their performance
degrades when applied outside the distribution of instances seen during
pretraining. Conversely, methods that require no offline cost at all
(LLM-to-Solution) rely on the zero-shot generalization of
general-purpose language models, which is known to degrade sharply on
larger or more structurally demanding instances~\cite{zhang2024}.
Neither category offers a model that is simultaneously trained
specifically for the instance at hand and free of external data
requirements. LM-GRASP is designed to occupy precisely this
intermediate position, trading the cross-instance generalization of a
pretrained policy for a learned constructor specialized, at no offline
cost, to the particular instance under optimization.

\section{The LM-GRASP Framework}
\label{sec:method}

This section describes the architecture and operational lifecycle of
LM-GRASP. We first provide a high-level overview and formalize the
algorithmic loop. Next, we detail Phase 1 (the bootstrap phase) and Phase 2 (the
iterative learn--infer--improve cycle). We then specify the algorithmic
role of the generative policy and the archive. We conclude with a
methodological discussion of computational trade-offs.

\subsection{Algorithmic Architecture Overview}

LM-GRASP replaces the hand-crafted randomized constructive phase of
GRASP with a learned parametric policy. The generation routine is parameterized by a
sequence model trained in an online imitation learning paradigm: the
generative policy is fit to the high-quality solutions stored in the
elite archive. No domain-specific heuristic rules or algebraic bounds are
encoded into the constructor. Instead, the pipeline interfaces with the
combinatorial domain solely through the objective function evaluator.

\begin{figure}[htbp]
\centering
\resizebox{\textwidth}{!}{%
\begin{tikzpicture}[
    node distance=1.5cm and 1.2cm,
    auto,
    core_style/.style={rectangle, draw=slate_blue!80, fill=slate_blue!8, text width=3.2cm, text centered, rounded corners=3pt, minimum height=1.3cm, font=\small\sffamily, thick},
    learn_style/.style={rectangle, draw=muted_amber!90, fill=muted_amber!10, text width=3.2cm, text centered, rounded corners=3pt, minimum height=1.3cm, font=\small\sffamily, thick},
    title_style/.style={font=\bfseries\sffamily\large, color=black!80, text centered},
    line/.style={draw=black!70, -{Stealth[scale=1.0]}, thick},
    loop_line/.style={draw=black!50, -{Stealth[scale=1.0]}, dashed, thick},
    slate_blue/.style={color=rgb,255:red=70;green=130;blue=180},
    muted_amber/.style={color=rgb,255:red=225;green=112;blue=85}
]

    \definecolor{slate_blue}{RGB}{65, 105, 225}
    \definecolor{muted_amber}{RGB}{214, 92, 0}

    \node[title_style] (titre_gpu) {GPU-GRASP (Classic)};
    
    \node[core_style, below=0.6cm of titre_gpu] (gpu_const) {
        \textbf{1. Greedy Construction}\\
        \scriptsize Probabilistic generation\\(Static $\alpha$-greedy)
    };
    
    \node[core_style, below=1.6cm of gpu_const] (gpu_ls) {
        \textbf{2. Local Search}\\
        \scriptsize Permutation improvement\\(Swap Best-Improvement)
    };
    
    \path[line] (gpu_const) -- node[midway, right, font=\scriptsize\itshape, color=black!60] {Initial sols.} (gpu_ls);
    
    \draw[line] (gpu_ls.west) -- ++(-0.4,0) |- node[pos=0.25, left, font=\scriptsize, align=center, color=black!70] {Repeat until\\budget\\exhausted} (gpu_const.west);

    \node[core_style, right=3.2cm of gpu_const] (lm_const) {
        \textbf{1. AI Construction}\\
        \scriptsize Autoregressive generation\\(\textbf{GPT-2-style Transformer})
    };
    
    \node[core_style, below=1.6cm of lm_const] (lm_ls) {
        \textbf{2. Local Search}\\
        \scriptsize Permutation improvement\\(\textbf{Expert Oracle})
    };

    \path[line] (lm_const) -- node[midway, right, font=\scriptsize\itshape, color=black!60] {Initial sols.} (lm_ls);
    
    \draw[loop_line] (lm_ls.west) -- ++(-0.4,0) |- node[pos=0.25, left, font=\scriptsize, color=black!80, align=center] {Inference\\Loop} (lm_const.west);

    \node[learn_style, right=1.2cm of lm_ls] (lm_archive) {
        \textbf{3. Elite Archive}\\
        \scriptsize Storage of high-quality\\trajectories
    };
    
    \node[learn_style, right=1.2cm of lm_const] (lm_train) {
        \textbf{4. Online Training}\\
        \scriptsize Behavioral cloning via\\gradient updates $\nabla$
    };

    \node[title_style] at ($(lm_const.north)!0.5!(lm_train.north) + (0,0.6cm)$) (titre_lm) {LM-GRASP (Proposed)};

    \path[line] (lm_ls) -- node[midway, below, font=\scriptsize\itshape, color=black!60] {Optimized paths} (lm_archive);
    \path[line] (lm_archive) -- node[midway, right, font=\scriptsize\itshape, color=black!60] {Updates} (lm_train);
    \path[line] (lm_train) -- node[midway, above, font=\scriptsize\itshape, color=black!60] {New policy} (lm_const);

    \begin{scope}[on background layer]
        \node[draw=slate_blue!30, fill=slate_blue!2, dashed, rounded corners=4pt, fit=(gpu_const) (gpu_ls), inner sep=0.3cm] (box_gpu) {};
        \node[draw=slate_blue!30, fill=slate_blue!2, dashed, rounded corners=4pt, fit=(lm_const) (lm_ls), inner sep=0.3cm] (box_lm_core) {};
        \node[draw=muted_amber!30, fill=muted_amber!2, dashed, rounded corners=4pt, fit=(lm_train) (lm_archive), inner sep=0.3cm] (box_lm_learn) {};
    \end{scope}
    
    \node[below=0.2cm of box_gpu.south, font=\small\bfseries\sffamily, color=slate_blue!80!black] {Standard Search Engine};
    \node[below=0.2cm of box_lm_core.south, font=\small\bfseries\sffamily, color=slate_blue!80!black] {Equivalent Core Engine};
    \node[below=0.2cm of box_lm_learn.south, font=\small\bfseries\sffamily, color=muted_amber!90!black] {Online Learning Loop};

\end{tikzpicture}%
}
\caption{Structural comparison between classic GPU-GRASP and the proposed LM-GRASP framework. The addition of the third column highlights the autonomous online learning loop, which interfaces seamlessly with the equivalent local search core.}
\label{fig-comparison-grasp-v3}
\end{figure}
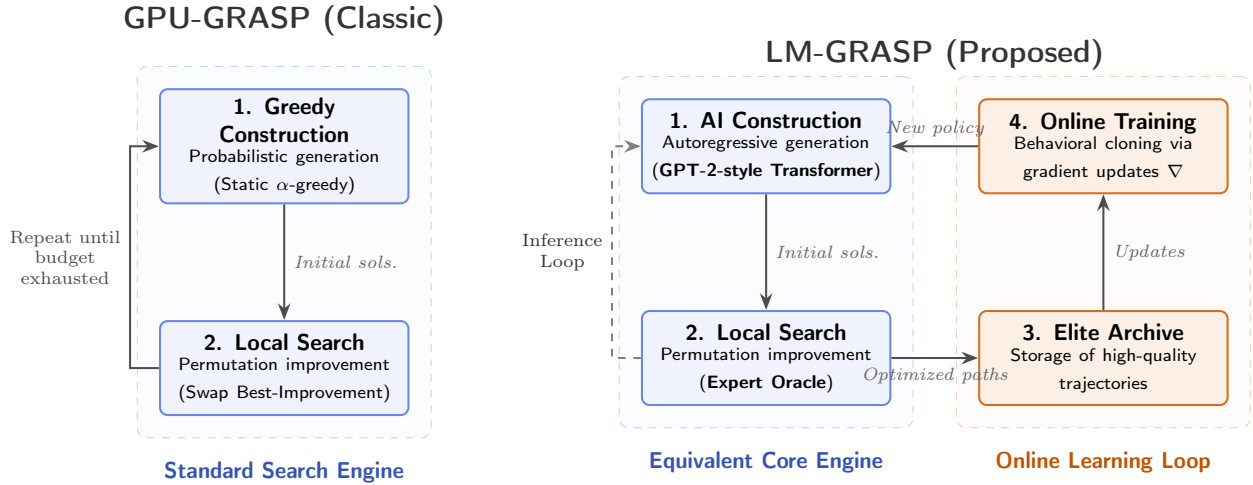

The optimization runtime consists of two distinct phases. Execution begins with an initial bootstrap
that seeds the elite archive (Line~1). This collection is followed by an iterative
learn--infer--improve loop (Lines~2--7) that runs until the computational budget is
completely exhausted. The lifecycle is formalized in Algorithm~\ref{alg:lmgrasp}
and illustrated in Figure~\ref{fig-comparison-grasp-v3}.

\begin{algorithm}[H]
\LinesNumbered
\caption{LM-GRASP Execution Loop}
\label{alg:lmgrasp}
\KwIn{Objective evaluator $f$, archive capacity $N_{\text{arch}}$,
      generative sample size $M$, wall-clock budget $T$}
\KwOut{Best solution found $x^*$}
$\mathcal{A} \leftarrow \textsc{Bootstrap}(N_{\text{arch}})$\;
\While{elapsed time $< T$}{
    $\theta \leftarrow \textsc{Train}(\pi_\theta, \mathcal{A})$\tcp*{Step (a): Online Policy Optimization}
    $\mathcal{S} \leftarrow \textsc{Infer}(\pi_\theta, M)$\tcp*{Step (b): Autoregressive Inference}
    $\mathcal{S} \leftarrow \textsc{Validate}(\mathcal{S})$\tcp*{Step (c): Feasibility \& Diversity Filtering}
    $\mathcal{A} \leftarrow \textsc{UpdateArchive}(\mathcal{A}, \mathcal{S}, N_{\text{arch}})$\tcp*{Step (d): Archive Eviction}
}
\Return $\arg\min_{x \in \mathcal{A}} f(x)$\;
\end{algorithm}

\subsection{Phase 1: Prior-Free Bootstrap Routine}

Before the policy can be trained, the elite archive $\mathcal{A}$ must
be seeded with an initial solution corpus (Line~1 of
Algorithm~\ref{alg:lmgrasp}). This step is achieved by generating
$N_{\text{boot}}$ permutations via Fisher-Yates shuffles. Each sequence is then evaluated
with the objective function $f$ and inserted into the archive. No
problem-specific knowledge is used at this stage. The resulting corpus
provides the initial training manifold for the first policy
optimization step.

\subsection{Phase 2: The Iterative Learn--Infer--Improve Loop}

Each generation $g$ of the iterative phase executes the four steps of
the \textbf{while} loop (Lines~2--7 of Algorithm~\ref{alg:lmgrasp}) in
sequence:

\paragraph{(a) Online Policy Optimization (Line~3).}
The policy $\pi_\theta$ is trained on the current contents of the
archive $\mathcal{A}$. Because the archive enforces a hard capacity of
$N_{\text{arch}}$ solutions with fitness-based retention, the repository serves
directly as the behavioral cloning corpus without additional
preprocessing. Parameters are optimized by minimizing the causal
cross-entropy loss:
$$ \mathcal{L}(\theta) = -\frac{1}{|\mathcal{A}|} \sum_{x \in \mathcal{A}} \sum_{t=1}^{n} \log p_\theta(x_t \mid x_{1:t-1}) $$
At generation 1, the network weights are initialized randomly. From generation 2
onward, training warm-starts from the checkpoint of the previous
generation. The warm-start strategy accelerates convergence on the incrementally updated
archive.

\paragraph{(b) Parallelized Autoregressive Inference (Line~4).}
The updated policy synthesizes a batch of $M$ candidate permutations
via ancestral sampling. Exploration is controlled by a top-$k$
truncation filter with $k = \lfloor \alpha \cdot n \rfloor$ and a
softmax temperature $\tau > 0$. A higher $\tau$ flattens the output
distribution to encourage diversity, while a lower $\tau$ concentrates probability mass
near the modes of the learned distribution.

\paragraph{(c) Feasibility and Diversity Filtering (Line~5).}
Each generated sequence is verified to be a valid, non-repeating
permutation of $\{0, \ldots, n-1\}$. Sequences that are invalid or
already present in the archive or current inference batch are immediately
discarded. The filter prevents premature mode collapse by ensuring that
only structurally distinct candidates move forward to the archive update step.

\paragraph{(d) Archive Eviction (Line~6).}
The unique, valid candidates $\mathcal{S}$ are inserted into the
archive. If the resulting size exceeds $N_{\text{arch}}$, the
highest-cost entries are removed. This eviction rule
maintains a direct correspondence between archive size and training
corpus size. Consequently, the updated archive can be used immediately in the
next generation without downsampling.

\subsection{Generative Policy and Archive: Algorithmic Roles}

The policy $\pi_\theta$ is an autoregressive sequence model whose
generative process is mapped onto the combinatorial construction step
of GRASP as follows:

\begin{itemize}
    \item A \textbf{token} corresponds directly to a \textbf{solution component}
    (such as a job index in a scheduling permutation).

    \item A \textbf{complete sequence} corresponds to a \textbf{feasible
    solution}.

    \item \textbf{Token selection} via the softmax distribution
    corresponds to the \textbf{stochastic element selection} from an RCL.

    \item \textbf{Sequence likelihood} serves as an implicit proxy for
    \textbf{solution quality}. Maximizing token log-likelihood encourages
    the policy to reproduce the structural patterns of elite solutions.
\end{itemize}

The vocabulary consists of the $n$ job indices augmented with
Begin-of-Sequence (\textsc{bos}) and End-of-Sequence (\textsc{eos})
tokens, giving a vocabulary size and context window of $n + 2$.

The archive $\mathcal{A}$ plays a dual algorithmic role: it is both the
elite pool from which the next bootstrap of training data is drawn
(Line~3) and the receptacle into which newly validated candidates are
written (Line~6). Telemetry collected at each generation tracks counts
of generated, invalid, duplicate, and novel solutions, as well as the
min, mean, and max objective values at three checkpoints per
iteration: before inference, after inference, and after the archive
update.

\subsection{Methodological Discussion}

The key difference between LM-GRASP and classical GRASP variants lies
in how the computational budget is allocated. GPU-GRASP maximizes the number of
independent random constructions per unit time via massive hardware
parallelism. In contrast, LM-GRASP invests part of the budget into policy
training (Line~3). This mechanism biases subsequent sampling (Line~4) toward high-quality regions of the
search space. We accept higher per-iteration overhead in exchange for
more focused candidate generation.

\section{Experiments}
\label{sec:experiments}

This section presents an empirical evaluation of LM-GRASP against
classical metaheuristic baselines. We describe the experimental setup,
report quantitative results on the Taillard benchmark suite, and
analyze convergence dynamics alongside intra-iteration policy behavior.

\subsection{Implementation Details}
\label{sec:implementation}

\textit{This subsection describes concrete implementation choices used
in our experiments; these are not part of the algorithm's specification
and can be substituted without altering Algorithm~\ref{alg:lmgrasp}.}

\paragraph{Policy network.} The policy $\pi_\theta$ is implemented as a
GPT-2-style decoder-only Transformer in PyTorch, with $L$
causally-masked self-attention layers, pre-LayerNorm blocks, learned
positional embeddings, and weight tying between the token embedding
matrix and the output projection head.

\paragraph{Training procedure.} Training uses the AdamW optimizer.
Weight decay is applied to parameter tensors of dimension $\geq 2$, and
the schedule combines cosine decay with linear warmup alongside
gradient norm clipping. An early-stopping monitor tracks validation
loss and terminates training if the improvement fails to exceed a
threshold $\varepsilon$ for $P$ consecutive evaluations, conserving the
wall-clock budget. Inference (Line~4) is vectorized into batched tensor
operations of size $B_{\text{inf}}$.

\paragraph{Archive storage.} The elite archive $\mathcal{A}$ is backed
by a SQLite database. Permutations are stored as text strings serving
as unique primary keys, paired with their objective cost; this layout
prevents duplicate entries and supports efficient SQL queries for
training-corpus extraction, fitness-distribution queries, and bounded
eviction (Line~6). Bootstrap solutions (Line~1) are written directly to
this database. A dedicated \texttt{inference\_stats} table records the
per-generation telemetry described above.

\subsection{Experimental Setup}

\subsubsection*{Computational Infrastructure}

We executed our experiments on two distinct platforms within the university cluster:

\begin{itemize}
    \item \textbf{Sequential CPU Baseline (Aion Cluster):} A BullSequana
    X2410 node equipped with two 64-core AMD EPYC 7H12 processors running at 2.6~GHz. 
    The configuration provides 128 physical cores total alongside 256~GB of DDR4 RAM running at 3200~MT/s.

    \item \textbf{GPU-Accelerated Frameworks (Iris Cluster):} A Dell
    PowerEdge C4140 node featuring dual 14-core Intel Xeon Gold 6132 CPUs at
    2.6~GHz. The platform includes 768~GB of DDR4 RAM at 2666~MT/s, paired with four
    NVIDIA Tesla V100-SXM2 GPUs interconnected via NVLink (300~GB/s).
    Both GPU-GRASP and LM-GRASP were bound to a single isolated V100
    instance to eliminate device-sharing interference.
\end{itemize}

\subsubsection*{Benchmark Protocol}

We evaluated all methods on the ten Taillard PFSP instances
\texttt{ta51}--\texttt{ta60}~\cite{taillard1993}. The problem set features 
$n = 50$ jobs and $m = 20$ machines ($Fm \| C_{\max}$), with a collective average
best-known makespan of 3708. Each method was run 10 times per instance
under an identical 5-hour wall-clock budget.

\paragraph{Choice of Instance Block.}
The Taillard benchmark spans configurations ranging from $20 \times 5$
to $500 \times 20$. Table~\ref{tab:taillard_optimums_connus} reports the
number of instances for which the exact optimum is known across the
full range as of June 2026, revealing that combinatorial difficulty is
not monotone in instance size: very large configurations such as
$500 \times 20$ or $200 \times 20$ are comparatively tractable, as their
fitness landscapes become highly homogeneous at scale, whereas the
genuine complexity frontier concentrates at the intermediate
$50 \times 20$ and $100 \times 20$ blocks, where the search landscape is
maximally rugged and a significant fraction of optima remain unknown.

The $50 \times 20$ block (\texttt{ta51}--\texttt{ta60}) is the first
configuration in the benchmark where exactly half of the absolute
optima are still open, making it the natural threshold at which the
structural advantage of a learned construction policy over a classical
greedy heuristic can be directly assessed: below this scale, the
computational efficiency of classical GRASP is unmatched, while at and
above it, the rugged search topology motivates the structural awareness
conferred by the Transformer policy.

\begin{table}[htbp]
\centering
\caption{Number of Taillard instances for which the exact optimum is
known (updated: June 2026). Configurations sharing the same count are
grouped; the two complexity-frontier blocks are shown individually.}
\label{tab:taillard_optimums_connus}
\begin{tabular}{lcc}
\toprule
\textbf{Configuration ($n \times m$)} & \textbf{Instance range} &
\textbf{Known optima / 10} \\
\midrule
$20 \times 5$, $20 \times 10$, $20 \times 20$, $50 \times 5$, $50 \times 10$ & ta001 -- ta050 & 10 / 10 (each) \\
\midrule
\textbf{$50 \times 20$} & \textbf{ta051 -- ta060} & \textbf{5 / 10} \\
\midrule
$100 \times 5$, $100 \times 10$ & ta061 -- ta080 & 10 / 10 (each) \\
\midrule
\textbf{$100 \times 20$} & \textbf{ta081 -- ta090} & \textbf{4 / 10} \\
\midrule
$200 \times 10$, $500 \times 20$ & ta091 -- ta100, ta111 -- ta120 & 10 / 10 (each) \\
$200 \times 20$ & ta101 -- ta110 &  9 / 10 \\
\midrule
\textbf{Total} & & \textbf{108 / 120} \\
\bottomrule
\end{tabular}
\end{table}

The $50 \times 20$ Taillard block is a recognized challenge for both
exact and approximate methods. Exact Branch-and-Bound requires
substantial computational resources on the designated instances. For example, instance
\texttt{ta56} required 25 days of computation (approximately 22
CPU-years) across 328 processors when first solved exactly by Mezmaz et
al.~\cite{mezmaz2007grid}. More recently, Gmys~\cite{gmys2022exactly}
solved several previously open instances using up to 384 V100 GPUs on
the Jean Zay supercomputer. 

For approximate methods, the fitness landscape is highly rugged~\cite{watson2003}, 
and standard local search operators stagnate readily. Previous GRASP-based approaches 
required coordinating 315 distinct hand-tuned configurations via a parallel
multi-core hyper-heuristic to remain competitive on the target block~\cite{alekseeva2017}. 
The severe difficulty motivates the use of a learned construction policy as an 
alternative to extensive manual tuning.

\subsubsection*{Evaluated Methods}

\begin{itemize}
    \item \textbf{CPU-GRASP:} Standard GRASP utilizing a hand-crafted
    constructive heuristic. The solver ranks jobs by their incremental makespan
    contribution and runs sequentially on the Aion CPU node.

    \item \textbf{GPU-GRASP:} A GPU-based GRASP variant
    using the same hand-crafted heuristic as CPU-GRASP. The script runs on the
    Iris V100 GPU. The performance difference between CPU-GRASP and
    GPU-GRASP isolates the effect of hardware acceleration under an
    identical construction rule.

    \item \textbf{LM-GRASP:} Our proposed framework, running on the
    same V100 GPU as GPU-GRASP. The construction policy is a
    decoder-only Transformer trained online via behavioral cloning on
    the top-$K$ solutions in a persistent archive. The system requires no
    domain-specific heuristic rules.
\end{itemize}

\subsubsection*{Model and Optimization Configuration}

\paragraph{Architecture.}
Following the implementation described in
Section~\ref{sec:implementation}, the Transformer policy uses
$L = 14$ attention layers, 12 attention heads, and an embedding
dimension of 768. We apply a dropout rate of 0.2 to the residual
connections, token embeddings, and attention weights.

\paragraph{Training.}
The archive size is fixed at $K = 5{,}000$ elite sequences. We use a
learning rate of $6 \times 10^{-4}$, a 500-step linear warmup, a weight
decay of 0.1, and a gradient norm clip of 1.0, with gradient
accumulation over 4 steps (effective batch size 64). Training runs for
at most 2,001 steps, with validation on a 10\% holdout every 500 steps;
early stopping triggers if validation loss does not improve by more
than $10^{-4}$ over 15 consecutive evaluations.

\paragraph{Inference.}
Generation uses a softmax temperature of $\tau = 1.3$ and top-$k$
filtering with $k = \lfloor 0.2 \times n \rfloor$, with candidates
sampled in batches of 512 ($B_{\text{inf}} = 512$), targeting a total
pool of $M = 10{,}000$ per generation.

\paragraph{Hyperparameter Selection.}
We established the architectural and optimization hyperparameters reported above 
empirically through a limited set of preliminary trials. We did not use exhaustive 
search procedures such as grid search or Bayesian optimization. The fact that competitive 
performance is achieved under an initial calibration underscores the robustness of the
LM-GRASP framework. A systematic sensitivity analysis would be particularly
costly for architectures of the current scale. The exploration falls outside the scope of
our introductory study and is identified as a natural direction for future work.

\subsection{Results and Analysis}

Tables~\ref{tab:results_best} and \ref{tab:results_avg} report the
best-found makespan across all runs alongside the per-run mean makespan.

\begin{table}[H]
\centering
\caption{Minimum makespan found across all runs and total run-level wins per method on Taillard instances \texttt{ta51}--\texttt{ta60}. \textbf{Bold} values indicate the best performer among tested methods.}
\label{tab:results_best}
\resizebox{\textwidth}{!}{
\begin{tabular}{l ccc cc ccc}
\toprule
                  & \multicolumn{3}{c}{\textbf{Minimum Makespan ($\min$)}} & \textbf{Best}  & \textbf{Is it} & \multicolumn{3}{c}{\textbf{Win Count ($\uparrow$)}} \\
\cmidrule(r){2-4}  \cmidrule(l){7-9}
\textbf{Instance} & \textbf{CPU-GRASP} & \textbf{GPU-GRASP} & \textbf{LM-GRASP} & \textbf{known} & \textbf{Opt.?} & \textbf{CPU-GRASP} & \textbf{GPU-GRASP} & \textbf{LM-GRASP} \\
\midrule
ta51 & 3950 & 3927 & \textbf{3898} & 3846 & $-$ & 0 & 0 & \textbf{10} \\
ta52 & 3806 & 3777 & \textbf{3751} & 3699 & $\checkmark$ & 0 & 0 & \textbf{10} \\
ta53 & 3758 & 3731 & \textbf{3693} & 3640 & $\checkmark$ & 0 & 0 & \textbf{10} \\
ta54 & 3823 & 3795 & \textbf{3769} & 3719 & $-$ & 0 & 0 & \textbf{10} \\
ta55 & 3719 & 3706 & \textbf{3662} & 3610 & $-$ & 0 & 0 & \textbf{10} \\
ta56 & 3775 & 3757 & \textbf{3725} & 3679 & $\checkmark$ & 0 & 0 & \textbf{10} \\
ta57 & 3810 & 3784 & \textbf{3760} & 3704 & $\checkmark$ & 0 & 0 & \textbf{10} \\
ta58 & 3812 & 3789 & \textbf{3753} & 3691 & $\checkmark$ & 0 & 0 & \textbf{10} \\
ta59 & 3856 & 3822 & \textbf{3797} & 3741 & $-$ & 0 & 0 & \textbf{10} \\
ta60 & 3832 & 3821 & \textbf{3811} & 3755 & $-$ & 0 & 0 & \textbf{10} \\
\midrule
\textbf{Average / Total} & 3814 & 3790 & \textbf{3761} & 3708 & & 0 & 0 & \textbf{100} \\
\bottomrule
\end{tabular}
}
\end{table}

\begin{table}[H]
\centering
\caption{Average of per-run best makespans per method on Taillard instances \texttt{ta51}--\texttt{ta60}. Values show $\text{mean} \pm \text{std. dev.}$ over 10 runs.}
\label{tab:results_avg}
\resizebox{\textwidth}{!}{
\begin{tabular}{lr ccc cc}
\toprule
 &  & \multicolumn{3}{c}{\textbf{Average Makespan Per Method}} & \multicolumn{2}{c}{\textbf{Pairwise Gap Evolution}} \\
\cmidrule(r){3-5} \cmidrule(l){6-7}
\textbf{Instance} & \textbf{\#Runs} & \textbf{CPU-GRASP} & \textbf{GPU-GRASP} & \textbf{LM-GRASP} & \textbf{CPU/GPU} & \textbf{GPU/LM} \\
\midrule
ta51 & 10 & \res{0}{3958.6}{5.8} & \res{0}{3933.3}{3.7} & \res{1}{3907.4}{4.1} & \res{0}{25.1}{6.2} & \res{0}{26.0}{4.9}  \\
ta52 & 10 & \res{0}{3820.1}{8.4} & \res{0}{3787.4}{8.0} & \res{1}{3758.8}{5.0} & \res{0}{37.3}{12.0} & \res{0}{27.8}{10.4}  \\
ta53 & 10 & \res{0}{3774.9}{9.1} & \res{0}{3740.7}{5.9} & \res{1}{3710.3}{9.7} & \res{0}{31.1}{12.5} & \res{0}{29.7}{12.9}  \\
ta54 & 10 & \res{0}{3831.2}{6.6} & \res{0}{3802.4}{3.8} & \res{1}{3776.6}{4.5} & \res{0}{28.0}{10.1} & \res{0}{26.0}{6.5}  \\
ta55 & 10 & \res{0}{3740.9}{13.4} & \res{0}{3714.6}{5.5} & \res{1}{3675.4}{7.3} & \res{0}{24.4}{16.0} & \res{0}{39.3}{10.8}  \\
ta56 & 10 & \res{0}{3789.5}{10.1} & \res{0}{3768.7}{7.2} & \res{1}{3739.7}{6.6} & \res{0}{21.3}{14.7} & \res{0}{28.3}{7.8}  \\
ta57 & 10 & \res{0}{3822.4}{7.2} & \res{0}{3797.6}{7.1} & \res{1}{3767.4}{4.0} & \res{0}{25.0}{12.7} & \res{0}{30.7}{8.1}  \\
ta58 & 10 & \res{0}{3825.6}{9.9} & \res{0}{3793.3}{4.3} & \res{1}{3761.7}{4.8} & \res{0}{32.4}{10.0} & \res{0}{31.0}{8.4}  \\
ta59 & 10 & \res{0}{3864.1}{5.5} & \res{0}{3836.1}{6.2} & \res{1}{3801.8}{3.2} & \res{0}{27.4}{5.9} & \res{0}{34.2}{6.7}  \\
ta60 & 10 & \res{0}{3849.0}{9.1} & \res{0}{3830.2}{6.5} & \res{1}{3818.8}{5.0} & \res{0}{19.6}{9.9} & \res{0}{11.2}{7.2}  \\
\midrule
\textbf{All instances} & 100 & \res{0}{3827.6}{8.5} & \res{0}{3800.4}{5.8} & \res{1}{3771.8}{5.4} & \res{0}{27.2}{11.0} & \res{0}{28.4}{8.4}  \\
\bottomrule
\end{tabular}
}
\end{table}

\begin{figure}[tbp]
    \centering
    \includegraphics[width=0.8\textwidth]{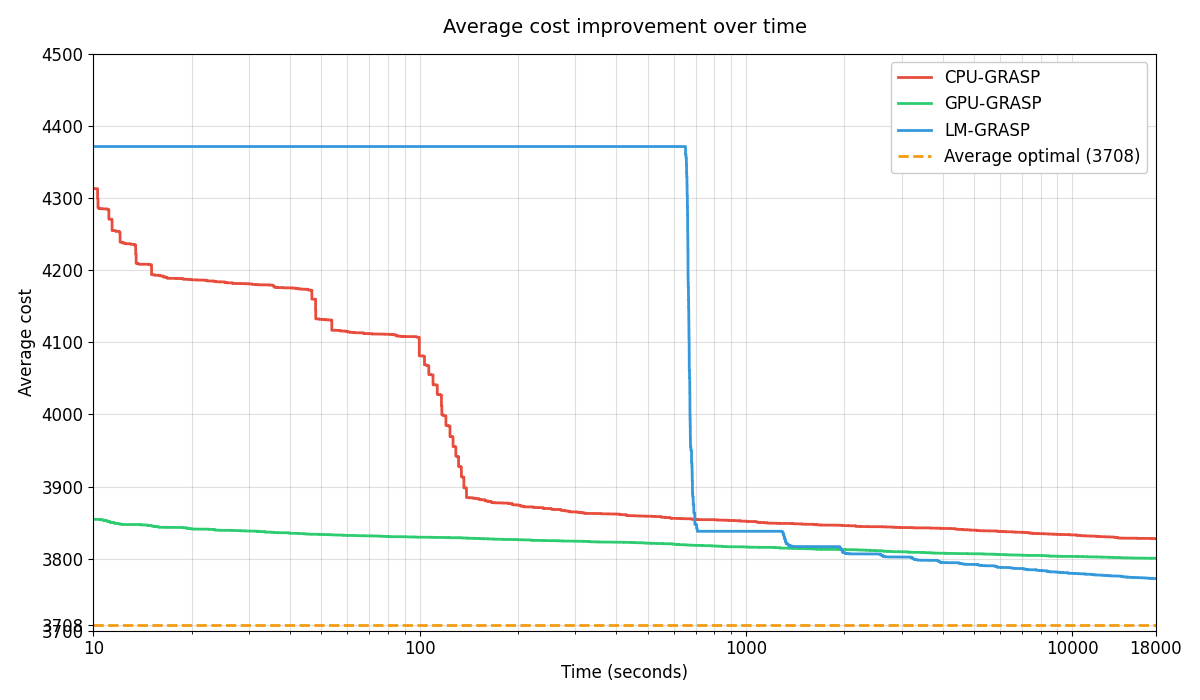}
    \caption{Average optimization trajectories over time (log scale)
    across \texttt{ta51}--\texttt{ta60} for all three methods. The
    dashed line marks the average best-known makespan (3708). GPU-GRASP
    converges to a plateau ($\approx$3800.4) within the first 30
    seconds. LM-GRASP spends approximately 660 seconds on bootstrapping
    and initial policy training, then improves steadily throughout the
    remaining budget, reaching a mean terminal makespan of 3771.8.}
    \label{fig:improvements}
\end{figure}

\begin{figure}[tbp]
    \centering
    \includegraphics[width=0.8\textwidth]{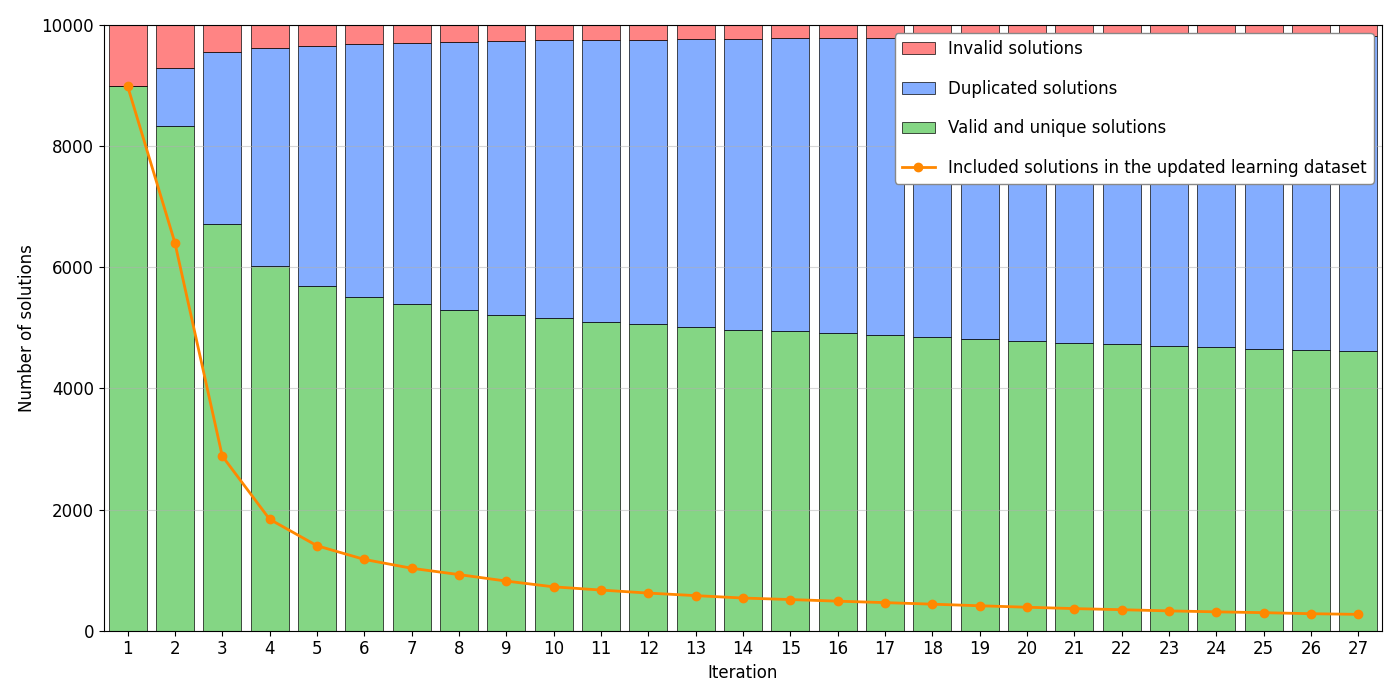}
    \caption{Breakdown of the 10,000-candidate batch by category across
    iterations. The proportion of novel, archive-improving candidates
    contracts from $\approx$8,986 in iteration 1 to $\approx$299 by
    iteration 25, while the share of duplicates increases correspondingly.
    This contraction reflects the natural exhaustion of promising,
    previously unexplored regions as the search converges.}
    \label{fig:occurrences}
\end{figure}

\begin{figure}[tbp]
    \centering
    \includegraphics[width=0.8\textwidth]{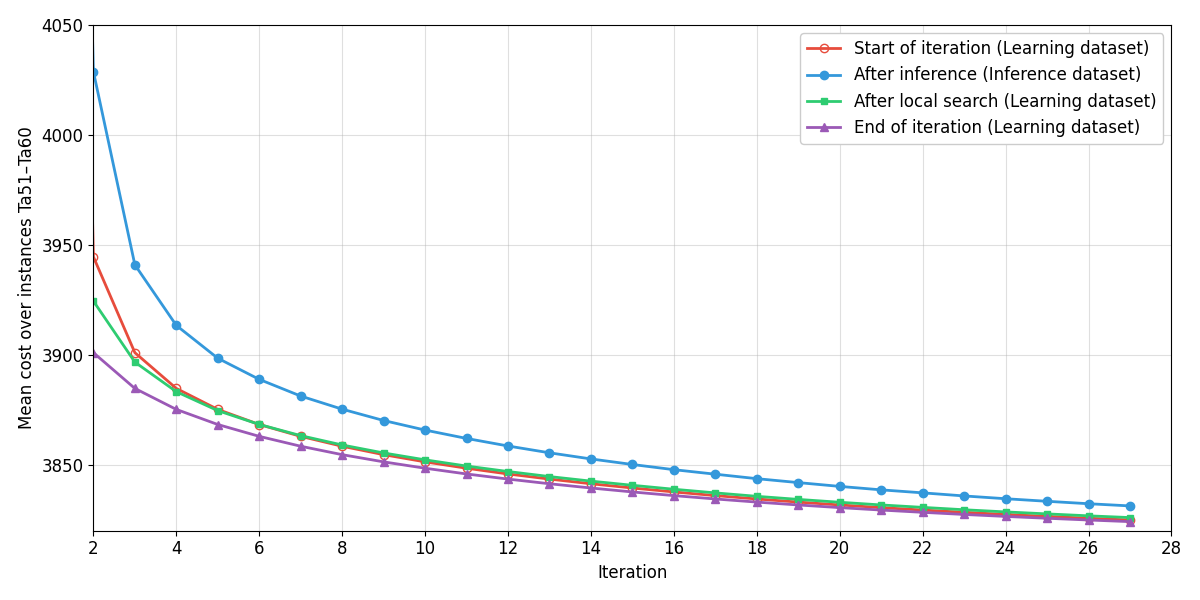}
    \caption{Intra-iteration makespan at four checkpoints per generation
    averaged across \texttt{ta51}--\texttt{ta60}: start of iteration
    (training corpus baseline), after inference, after local search, and
    end of iteration (archived). The monotonic decrease of end-of-iteration values
    from $\approx$3900 to $\approx$3824 confirms stable convergence of
    the optimization loop.}
    \label{fig:progress}
\end{figure}

\paragraph{Comparative Performance.}
LM-GRASP achieves a mean makespan of $3771.8 \pm 5.4$ across all
instances and runs. In comparison, GPU-GRASP scores $3800.4 \pm 5.8$ and
CPU-GRASP scores $3827.6 \pm 8.5$. LM-GRASP emerges as the best-performing method
on all ten instances (Table~\ref{tab:results_best}) and across all 100
individual runs (Table~\ref{tab:results_avg}).

Our three-way design decouples two performance components. The
\textbf{hardware acceleration delta} (CPU-GRASP vs.\ GPU-GRASP) averages 
27.2 units. Because both variants use the exact same construction
rule, the throughput gap reflects pure parallelization capacity. The
\textbf{algorithmic learning delta} (GPU-GRASP vs.\ LM-GRASP) averages 28.4
units. Because both variants run on identical V100 hardware,
the resulting improvement is directly attributable to the learned construction policy. 

The two deltas are of similar magnitude, showing a difference of only 1.2 units, 
with standard deviations of $\pm 11.0$ and $\pm 8.4$ respectively. The proximity
suggests that the algorithmic contribution is at least comparable to the hardware 
contribution in our testing framework. We note that the comparison is specific to the evaluated 
instances and budget. 

\paragraph{Convergence Dynamics.}
Figure~\ref{fig:improvements} illustrates divergent convergence
profiles: GPU-GRASP quickly exhausts its hardware-parallel exploration
and plateaus, whereas LM-GRASP follows a two-phase trajectory, investing
early budget in bootstrapping and policy training before improving
steadily thereafter. Crucially, the LM-GRASP trajectory has not
flattened at budget exhaustion, suggesting that additional compute time
would yield further gains.

\paragraph{Intra-Iteration Policy Generalization.}
Figure~\ref{fig:progress} shows that, from iteration 5 onward, the
policy generates solutions of lower mean makespan than its own training
corpus, indicating that it does not merely reproduce archived solutions
but generalizes to produce novel sequences of higher quality.

\paragraph{Convergence of the Generative Process.}
Figure~\ref{fig:occurrences} documents a progressive contraction in the
number of archive-improving candidates across iterations. Rather than a
pathological failure mode, this is the natural signature of
convergence: as the search identifies and archives the highest-quality
regions of the solution space, fewer unexplored improvements remain
available, consistent with an optimization loop approaching the limits
of the accessible search space under the current budget.

These results support our central claim: a construction policy learned
online via behavioral cloning can replace a hand-crafted greedy
heuristic and produce better solutions under an equivalent resource
budget, with the local search oracle alone providing a sufficient
training signal and no domain-specific feature encoding. Compared to
prior work on this benchmark block, LM-GRASP achieves strong
performance through a single neural network, avoiding the coordinated
selection among hundreds of hand-tuned configurations required by
previous GRASP-based approaches~\cite{alekseeva2017}.

The two methods represent different strategies for using a fixed
budget: GPU-GRASP prioritizes breadth by maximizing the number of
independent constructions, while LM-GRASP prioritizes depth by
investing in a generative model that biases sampling toward
high-quality regions. Because the anytime trajectories show no plateau
at budget exhaustion (Figure~\ref{fig:improvements}), extending the
search horizon -- through longer budgets or continuous policy updates
-- represents a natural path to further improvement.

\section{Conclusion}
\label{sec:conclusion}

We introduced LM-GRASP, a hybrid metaheuristic that replaces the
randomized constructive phase of GRASP with a policy learned online via
behavioral cloning. A local search procedure serves as the expert
oracle, a decoder-only Transformer serves as the constructive policy,
and the training corpus is built from elite solutions discovered during
the search itself, eliminating the need for domain-specific feature
engineering or external training data.

On the Taillard \texttt{ta51}--\texttt{ta60} instances under a 5-hour
budget, LM-GRASP outperforms both CPU-GRASP and GPU-GRASP on every
instance and every run (Section~\ref{sec:experiments}). Our
three-tiered hardware design isolates a hardware acceleration delta
(CPU-GRASP vs.\ GPU-GRASP) from an algorithmic learning delta
(GPU-GRASP vs.\ LM-GRASP); the two are of comparable magnitude, which
suggests that a learned construction policy is at least as effective as
hardware acceleration at improving solution quality. Intra-iteration
analysis further shows that, from iteration 5 onward, the policy
generates solutions of higher quality than its own training corpus,
confirming that the model generalizes beyond the archived demonstrations
rather than reproducing them.

The $50 \times 20$ Taillard block was chosen because it sits at the
complexity frontier of the benchmark, where classical greedy
construction begins to show its structural limits. Consistent with the
No Free Lunch theorem~\cite{wolpert2002no}, LM-GRASP is not
intended as a universal replacement for GRASP: for smaller or
lower-dimensional instances, classical GRASP remains the more efficient
choice, and our tool is best viewed as a targeted alternative for
problem regimes complex enough to amortize the cost of online policy
training. Consistent with this reading, the anytime trajectories show
no plateau at budget exhaustion, and the rate of archive-improving
candidates declines gradually rather than collapsing outright,
indicating that the search has not yet saturated the accessible
horizon.

Several extensions follow naturally from our study:

\begin{itemize}
    \item \textbf{Cross-Domain Validation:} The pipeline requires only
    an integer sequence representation paired with an objective
    evaluator, so it applies directly to other combinatorial problems.
    Evaluating LM-GRASP on the TSP, the Quadratic Assignment Problem, and
    bin-packing will test whether the observed advantages transfer
    across problem classes, and help delineate the complexity threshold
    below which classical GRASP remains preferable.

    \item \textbf{Continuous Policy Updates:} Replacing batch retraining
    with continuous online gradient updates as elite solutions enter the
    archive would reduce the latency between discovery and learning,
    potentially accelerating convergence and extending the effective
    search horizon within a fixed budget.

    \item \textbf{Hyperparameter Optimization:} The architectural and
    inference parameters used in our evaluation were set empirically
    through preliminary trials. A systematic sensitivity analysis,
    particularly of the temperature schedule and archive capacity, is a
    necessary next step to characterize the operating envelope of
    LM-GRASP.
\end{itemize}

Our work demonstrates that a construction policy learned online from
self-generated expert demonstrations can replace a hand-crafted greedy
heuristic and produce highly competitive solutions under an equivalent
computational budget, without any problem-specific feature design.

\section*{Code Availability}

The implementation of LM-GRASP is available as part of the COLM
platform~\cite{mezmaz2026}, at \url{https://gitlab.com/uniluxembourg/snt/pcog/colm}.

\section*{Acknowledgment}

The experiments presented in this paper were carried
out using the HPC facilities of the University of Luxembourg
~\cite{varrette2022} {\small -- see \url{https://hpc.uni.lu}}

\bibliographystyle{plainnat}
\bibliography{references}

@article{alekseeva2017,
  author  = {Alekseeva, E. and Mezmaz, M. and Tuyttens, D. and Melab, N.},
  title   = {Parallel multi-core hyper-heuristic {GRASP} to solve permutation flow-shop problem},
  journal = {Concurrency and Computation: Practice and Experience},
  year    = {2017},
  volume  = {29},
  number  = {9},
  pages   = {e3835}
}

@inproceedings{chen2019,
  author    = {Chen, X. and Tian, Y.},
  title     = {Learning to perform local rewriting for combinatorial optimization},
  booktitle = {Advances in Neural Information Processing Systems},
  volume    = {32},
  publisher = {Curran Associates, Inc.},
  year      = {2019}
}

@misc{chin2024,
  author = {Chin, S.~J. and Winkenbach, M. and Srivastava, A.},
  title  = {Learning to deliver: A foundation model for the {M}ontreal capacitated vehicle routing problem},
  year   = {2024},
  note   = {arXiv preprint arXiv:2403.00026}
}

@misc{daros2025,
  author = {Da Ros, F. and Soprano, M. and Di Gaspero, L. and Roitero, K.},
  title  = {Large language models for combinatorial optimization: A systematic review},
  year   = {2025},
  note   = {arXiv preprint arXiv:2507.03637}
}

@article{elhenawy2024,
  author  = {Elhenawy, M. and others},
  title   = {Visual reasoning and multi-agent approach in multimodal large language models: Solving {TSP} and {MTSP} combinatorial challenges},
  journal = {Machine Learning and Knowledge Extraction},
  year    = {2024},
  volume  = {6},
  number  = {3},
  pages   = {1894--1920}
}

@article{feo1995,
  author  = {Feo, T.~A. and Resende, M.~G.~C.},
  title   = {Greedy randomized adaptive search procedures},
  journal = {Journal of Global Optimization},
  year    = {1995},
  volume  = {6},
  number  = {2},
  pages   = {109--133}
}

@article{garey1976,
  author  = {Garey, M.~R. and Johnson, D.~S. and Sethi, R.},
  title   = {The complexity of flowshop and jobshop scheduling},
  journal = {Mathematics of Operations Research},
  year    = {1976},
  volume  = {1},
  number  = {2},
  pages   = {117--129}
}

@inproceedings{gasse2019,
  author    = {Gasse, M. and Ch{\'e}telat, D. and Ferroni, N. and Charlin, L. and Lodi, A.},
  title     = {Exact combinatorial optimization with graph convolutional neural networks},
  booktitle = {Advances in Neural Information Processing Systems},
  volume    = {32},
  year      = {2019}
}

@article{gmys2022exactly,
  author  = {Gmys, J.},
  title   = {Exactly solving hard permutation flowshop scheduling problems on peta-scale {GPU}-accelerated supercomputers},
  journal = {INFORMS Journal on Computing},
  year    = {2022},
  volume  = {34},
  number  = {5},
  pages   = {2448--2466}
}

@article{held1962dynamic,
  author  = {Held, M. and Karp, R.~M.},
  title   = {A dynamic programming approach to sequencing problems},
  journal = {Journal of the Society for Industrial and Applied Mathematics},
  year    = {1962},
  volume  = {10},
  number  = {1},
  pages   = {196--210}
}

@article{hopfield1985,
  author  = {Hopfield, J.~J. and Tank, D.~W.},
  title   = {``{N}eural'' computation of decisions in optimization problems},
  journal = {Biological Cybernetics},
  year    = {1985},
  volume  = {52},
  number  = {3},
  pages   = {141--152}
}

@inproceedings{huang2025,
  author    = {Huang, X. and Shen, Q. and Hu, Y. and Gao, A. and Wang, B.},
  title     = {{LLM}s for mathematical modeling: Towards bridging the gap between natural and mathematical languages},
  booktitle = {Findings of ACL: NAACL 2025},
  pages     = {2678--2710},
  year      = {2025}
}

@article{hussein2017,
  author  = {Hussein, A. and Gaber, M.~M. and Elyan, E. and Jayne, C.},
  title   = {Imitation learning: A survey of learning methods},
  journal = {ACM Computing Surveys},
  year    = {2017},
  volume  = {50},
  number  = {2},
  pages   = {1--35}
}

@inproceedings{kool2019,
  author    = {Kool, W. and van Hoof, H. and Welling, M.},
  title     = {Attention, learn to solve routing problems!},
  booktitle = {International Conference on Learning Representations (ICLR)},
  year      = {2019}
}

@inproceedings{kwon2020,
  author    = {Kwon, Y.-D. and Choo, J. and Kim, B. and Yoon, I. and Gwon, Y. and Min, S.},
  title     = {{POMO}: Policy optimization with multiple optima for reinforcement learning},
  booktitle = {Advances in Neural Information Processing Systems (NeurIPS)},
  pages     = {21188--21198},
  year      = {2020}
}

@misc{liu2024,
  author = {Liu, F. and Tong, X. and Yuan, M. and Lin, X. and Luo, F. and Wang, Z. and Lu, Z. and Zhang, Q.},
  title  = {Evolution of heuristics: Towards efficient automatic algorithm design using large language model},
  year   = {2024},
  note   = {arXiv preprint arXiv:2401.02051}
}

@inproceedings{martinek2024,
  author    = {Martinek, A. and Lukasik, S. and Gandomi, A.~H.},
  title     = {Large language models as tuning agents of metaheuristics},
  booktitle = {ESANN 2024},
  year      = {2024}
}

@inproceedings{mezmaz2007grid,
  author    = {Mezmaz, M. and Melab, N. and Talbi, E.-G.},
  title     = {A grid-enabled branch and bound algorithm for solving challenging combinatorial optimization problems},
  booktitle = {Proceedings of the 2007 IEEE International Parallel and Distributed Processing Symposium (IPDPS)},
  pages     = {1--9},
  year      = {2007}
}

@misc{mezmaz2026,
  author       = {Mezmaz, M.},
  title        = {{COLM}: A platform for integrating language models in combinatorial optimization},
  year         = {2026},
  howpublished = {\url{https://gitlab.com/uniluxembourg/snt/pcog/colm}}
}

@article{pomerleau1991,
  author  = {Pomerleau, D.~A.},
  title   = {Efficient training of artificial neural networks for autonomous navigation},
  journal = {Neural Computation},
  year    = {1991},
  volume  = {3},
  number  = {1},
  pages   = {88--97}
}

@article{ravetti2012,
  author  = {Ravetti, M.~G. and Riveros, C. and Mendes, A. and Resende, M.~G. and Pardalos, P.~M.},
  title   = {Parallel hybrid heuristics for the permutation flow shop problem},
  journal = {Annals of Operations Research},
  year    = {2012},
  volume  = {199},
  number  = {1},
  pages   = {269--284}
}

@inproceedings{ross2011,
  author    = {Ross, S. and Gordon, G. and Bagnell, D.},
  title     = {A reduction of imitation learning and structured prediction to no-regret online learning},
  booktitle = {Proceedings of the Fourteenth International Conference on Artificial Intelligence and Statistics},
  pages     = {627--635},
  publisher = {JMLR Workshop and Conference Proceedings},
  year      = {2011}
}

@article{taillard1993,
  author  = {Taillard, E.},
  title   = {Benchmarks for basic scheduling problems},
  journal = {European Journal of Operational Research},
  year    = {1993},
  volume  = {64},
  number  = {2},
  pages   = {278--285}
}

@inproceedings{varrette2022,
  author    = {Varrette, S. and Cartiaux, H. and Peter, S. and Kieffer, E. and Valette, T. and Olloh, A.},
  title     = {Management of an Academic {HPC} \& Research Computing Facility: The {ULHPC} Experience 2.0},
  booktitle = {Proceedings of the 6th ACM High Performance Computing and Cluster Technologies Conference (HPCCT 2022)},
  address   = {Fuzhou, China},
  publisher = {Association for Computing Machinery (ACM)},
  note      = {ISBN 978-1-4503-9664-6},
  year      = {2022}
}

@inproceedings{vaswani2017,
  author    = {Vaswani, A. and Shazeer, N. and Parmar, N. and Uszkoreit, J. and Jones, L. and Gomez, A.~N. and Kaiser, {\L}. and Polosukhin, I.},
  title     = {Attention is all you need},
  booktitle = {Advances in Neural Information Processing Systems},
  volume    = {30},
  year      = {2017}
}

@inproceedings{vinyals2015,
  author    = {Vinyals, O. and Fortunato, M. and Jaitly, N.},
  title     = {Pointer networks},
  booktitle = {Advances in Neural Information Processing Systems (NeurIPS)},
  volume    = {28},
  year      = {2015}
}

@article{watson2003,
  author  = {Watson, J.~P. and Beck, J.~C. and Howe, A.~E. and Whitley, L.~D.},
  title   = {Problem difficulty for permutation flow-shop scheduling},
  journal = {Artificial Intelligence},
  year    = {2003},
  volume  = {149},
  number  = {2},
  pages   = {199--231}
}

@article{wolpert2002no,
  author  = {Wolpert, D.~H. and Macready, W.~G.},
  title   = {No free lunch theorems for optimization},
  journal = {IEEE Transactions on Evolutionary Computation},
  year    = {2002},
  volume  = {1},
  number  = {1},
  pages   = {67--82}
}

@inproceedings{zhang2024,
  author    = {Zhang, T. and Yuan, J. and Avestimehr, S.},
  title     = {Revisiting {OPRO}: The limitations of small-scale {LLM}s as optimizers},
  booktitle = {Findings of ACL 2024},
  pages     = {1727--1735},
  year      = {2024}
}

\end{document}